\documentclass{article}

 \usepackage{graphicx}

 \usepackage[main, final]{neurips_2025}

\usepackage{listings}
\usepackage[utf8]{inputenc} 
\usepackage[T1]{fontenc}    
\usepackage{hyperref}       
\usepackage{url}            
\usepackage{booktabs}       
\usepackage{amsfonts}       
\usepackage{nicefrac}       
\usepackage{microtype}      
\usepackage{xcolor}         
\usepackage{amsmath}        
\usepackage{algorithm}      
\usepackage{algorithmic}    
\usepackage{subcaption}

\title{Evaluation of Vision-LLMs in Surveillance Video}

\author{%
  Pascal Benschop \\
  Department of Computer Science \\
  Delft University of Technology \\
  Delft, Netherlands \\
  \texttt{P.Benschop@tudelft.nl} \\
  \And
  Cristian Meo \\
  LatentWorlds AI \\
  Delft University of Technology \\
  Delft, Netherlands \\
  \texttt{C.Meo@tudelft.nl}
  \And
  Justin Dauwels \\
  Department of Computer Science \\
  Delft University of Technology \\
  Delft, Netherlands \\
  \texttt{J.H.G.Dauwels@tudelft.nl}
  \And
  Jelte P. Mense\\
  National Policelab AI \& Model-Driven Decisions Lab \\
  Delft University of Technology \\
  Delft, Netherlands \\
  \texttt{j.p.mense@tudelft.nl}
}

\begin{document}

\maketitle

\begin{abstract}
The widespread use of cameras in our society has created an overwhelming amount of video data, far exceeding the capacity for human monitoring. This presents a critical challenge for public safety and security, as the timely detection of anomalous or criminal events is crucial for effective response and prevention. The ability for an embodied agent to recognize unexpected events is fundamentally tied to its capacity for spatial reasoning. This paper investigates the spatial reasoning of vision-language models (VLMs) by framing anomalous action recognition as a zero-shot, language-grounded task, addressing the embodied perception challenge of interpreting dynamic 3D scenes from sparse 2D video. Specifically, we investigate whether small, pre-trained vision--LLMs can act as \emph{spatially-grounded}, zero-shot anomaly detectors by converting video into text descriptions and scoring labels via textual entailment. We evaluate four open models on UCF-Crime and RWF-2000 under prompting and privacy-preserving conditions. Few-shot exemplars can improve accuracy for some models, but may increase false positives, and privacy filters---especially full-body GAN transforms---introduce inconsistencies that degrade accuracy. These results chart where current vision--LLMs succeed (simple, spatially salient events) and where they falter (noisy spatial cues, identity obfuscation). Looking forward, we outline concrete paths to strengthen spatial grounding without task-specific training: structure-aware prompts, lightweight spatial memory across clips, scene-graph or 3D-pose priors during description, and privacy methods that preserve action-relevant geometry. This positions zero-shot, language-grounded pipelines as adaptable building blocks for embodied, real-world video understanding. Our implementation for evaluating VLMs is publicly available at: \url{https://github.com/pascalbenschopTU/VLLM_AnomalyRecognition}

\end{abstract}


\section{Introduction}

Zero-shot action recognition has emerged as a promising approach for labeling previously unseen video data, which is especially relevant for applications such as surveillance and anomaly detection. Recent advances in large vision-language models have demonstrated impressive performance on standard action recognition tasks~\cite{VideoLLaMA3, gemma3, nvila, Qwen2.5-VL}, largely by leveraging the transfer capabilities of pre-trained language models. However, there is little to no research validating whether these models can generalize to \textit{anomalous} action recognition—a domain characterized by rare, atypical, or criminal events that are often underrepresented or entirely absent from standard training datasets.
For high-stakes settings like security or forensic analysis, detecting anomalous actions automatically could help in settings where the amount of cameras far exceeds the amount of operators. To do this reliably, substantial training and testing data are required, and this data should be anonymized with privacy filters. Public datasets for anomalous action recognition, such as UCF-Crime~\cite{UCFCrime}, XD-Violence~\cite{XD_violence}, and RWF-2000~\cite{RWF2000}, are limited in scope, size, and label diversity. As a result, models trained or evaluated only on these datasets may not generalize well to new types of anomalies, and some relevant behaviors may not be covered at all.

In this work, we systematically evaluate the zero-shot capabilities of several state-of-the-art small ($\leq$8B params) vision-LLMs across a range of anomalous action recognition benchmarks and experimental conditions. We focus on two central research questions:
\textbf{RQ1:} How can current vision-LLMs be adapted to recognize criminal or anomalous actions in a zero-shot setting, and what role does few-shot prompting play in shaping their predictions? \textbf{RQ2:} How do privacy-preserving transformations (e.g., blurring or appearance changes) affect the ability of vision-LLMs to detect and classify anomalous events?

To answer these questions, we design controlled experiments using two benchmark datasets (UCF-Crime and RWF-2000) and four representative vision-LLMs, under multiple prompting strategies: unguided, guided, with privacy filtering, and with few-shot prompting. Our analysis includes both quantitative metrics and fault analysis to identify typical model errors and failure cases.
Our findings provide new insight into the limits and biases of current vision-LLMs for real-world anomalous action recognition, with practical recommendations for improving robustness—including strategies for frame sampling, prompt engineering, and privacy-aware preprocessing. This work aims to inform the design of safer, more effective video understanding systems for critical domains.

\section{Related Work}
\paragraph{Training-based Anomalous Action Recognition (UCF-Crime/XD-Violence).}
UCF-Crime introduced weakly supervised MIL over long, untrimmed videos~\cite{UCFCrime}, while XD-Violence added large-scale multimodal (audio–visual) supervision~\cite{XD_violence}.
Recent systems emphasize efficiency/real-time deployment (e.g., REWARD~\cite{REWARD} and AnomalyCLIP~\cite{AnomalyCLIP}) and structured reasoning via multimodal GNNs with mission-specific knowledge graphs (MissionGNN)~\cite{misionGNN}.
These methods, while not zero-shot, set competitive supervised or weakly supervised baselines for anomalous recognition on UCF-Crime/XD-Violence.
\paragraph{Vision–Language Models (VLMs) and Zero-Shot Anomaly Recognition.}
VLMs have been adapted to surveillance in several ways.
AnomalyCLIP reshapes CLIP’s latent space and learns a classifier for anomaly classes, achieving strong recognition but not strictly zero-shot~\cite{AnomalyCLIP}.
Training-free pipelines like LAVAD caption frames and prompt an LLM to aggregate anomalies temporally~\cite{lavad, meo2024object}, and Holmes-VAD instruction-tunes a multimodal LLM for interpretable VAD within a supervised pipeline~\cite{holmesVAD}.
Other caption-driven works (e.g., TEVAD) leverage text to improve anomaly scoring~\cite{tevad}, while open-vocabulary VAD explores generalization beyond closed sets~\cite{openvocab}.
Recent open vision-LLMs (e.g., NVILA~\cite{nvila} and VideoLLaMA3~\cite{VideoLLaMA3}) provide stronger video understanding backbones for zero-shot probing, but have not been systematically evaluated for privacy-robust anomaly recognition on UCF-Crime or RWF-2000~\cite{RWF2000}.

\section{Methodology}
Traditional approaches to video anomaly detection often rely on supervised learning, requiring extensive datasets with meticulously annotated event boundaries and classes~\cite{UCFCrime,REWARD,AnomalyCLIP}. This paradigm is costly, scales poorly, and fundamentally struggles to recognize novel or rare anomalies not present in the training data. In contrast to these data-hungry methods, our goal is to develop a framework that can identify and classify anomalous events in a zero-shot, training-free manner. We seek to leverage the powerful semantic reasoning and world knowledge embedded \cite{meo2024masked} within large, pre-trained vision-LLMs. The core motivation is to reframe anomaly classification not as a pixel-to-label mapping problem, but as a language-grounded reasoning task. By prompting a model to first describe a video's content in natural language and then using a separate text classifier to evaluate this description against human-readable labels, we can create a flexible and adaptable system that requires no task-specific fine-tuning or parameter updates.

\subsection{Derivation}
We formally derive our training-free anomaly classification framework. Let a video be represented as a sequence of RGB frames $X=(x_t)_{t=1}^{T}$, where each frame $x_t \in \{0, \dots, 255\}^{H \times W \times 3}$. The task is to assign a label from a predefined set of human-readable anomaly classes, $\mathcal{L}=\{\ell_1, \ell_2, \dots, \ell_C\}$. The process is composed of two main steps:

\textbf{Textual Description Generation:}
The central component is a vision-LLM, $F_{\theta}$, with frozen parameters $\theta$. This model processes the input video $X$ to generate a concise, descriptive text string, $t$. This generation is conditioned on the visual input and an optional textual prompt, $p$, which provides context for the task. The output description is sampled from the model's predictive distribution:
$$t \sim p_{\theta}(\cdot \mid X, p)$$

\textbf{Zero-Shot Classification via NLI:}
The generated text $t$ is then evaluated by a pre-trained, frozen Natural Language Inference (NLI) classifier, $g_{\phi}$, with parameters $\phi$. We cast the classification as a zero-shot textual entailment problem. For each candidate label $\ell_j \in \mathcal{L}$, the classifier computes a score, $s_j$, that quantifies the degree to which the generated description $t$ logically entails the label $\ell_j$. Formally, $s_j = g_{\phi}(t, \ell_j) \quad \forall \ell_j \in \mathcal{L}$. The final classification, $\hat{\ell}$, is the anomaly label that receives the highest entailment score, thus representing the most plausible description of the event in the video. This entire pipeline, from raw video frames to a final class label, operates without any gradient-based updates to either the vision-LLM ($F_{\theta}$) or the NLI classifier ($g_{\phi}$).
\subsection{Practical Implications}
The proposed framework has several practical advantages over traditional supervised models.

\textbf{True Zero-Shot Flexibility:}
Its primary strength is the ability to classify anomalies it has never been trained on. New anomaly types can be detected simply by adding a corresponding text label to the set $\mathcal{L}$, without any modification to the models, making the system adaptable to evolving requirements.

\textbf{Modular and Upgradable:}
The architecture is inherently modular. The vision-LLM and the NLI classifier are decoupled components that can be independently updated or replaced. For instance, a more advanced vision-LLM can be integrated into the pipeline to improve visual understanding without altering the classification module, facilitating straightforward performance enhancements.

\begin{figure}[b]
    \centering
    \includegraphics[width=\linewidth]{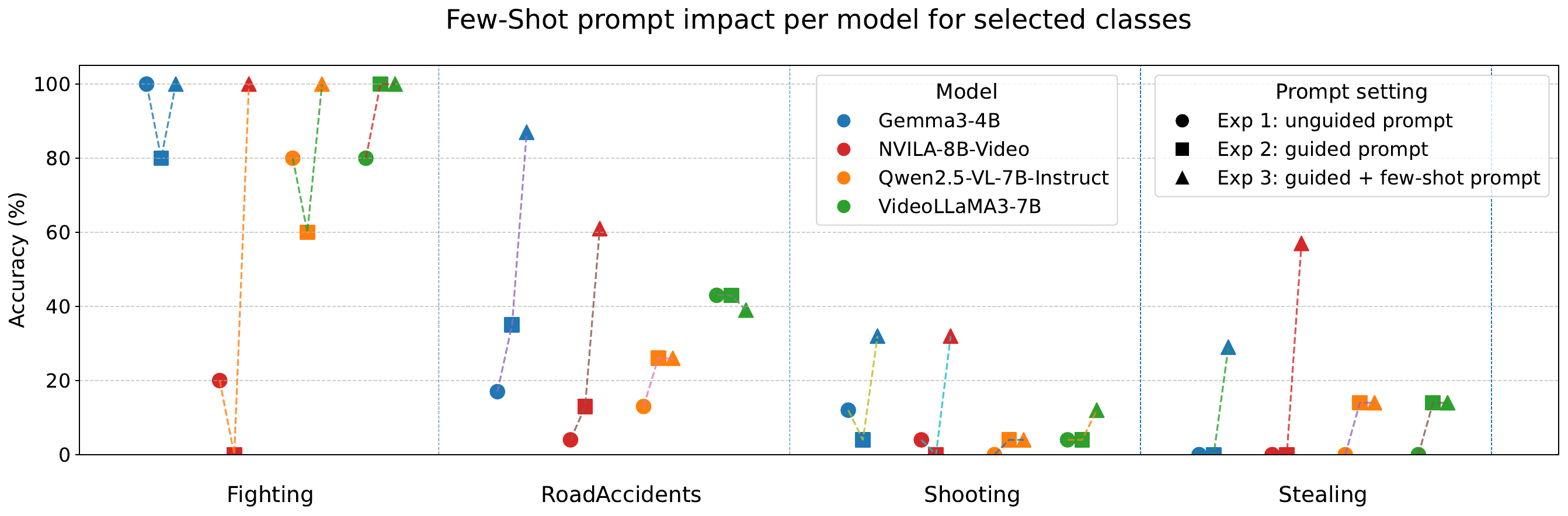}
    \caption{Few-shot prompting impact on models accuracy level.}
    \label{fig:fewshot_results}
\end{figure}
\section{Experimental Setup}

To assess the performance of our training-free framework, we evaluate its zero-shot anomalous action classification across models, prompting regimes, and privacy-preserving conditions. We validate on two standard video anomaly detection benchmarks using their canonical splits: UCF-Crime~\cite{UCFCrime}, which contains 13 anomaly classes plus normal videos, and RWF-2000~\cite{RWF2000}, which comprises real-world videos labeled as normal or fighting. We evaluate four open-source vision-LLMs off the shelf (no additional training): Gemma-3 (4B)~\cite{gemma3}, Qwen-2.5-VL-7B-Instruct~\cite{Qwen2.5-VL}, VideoLLaMA-3-7B~\cite{VideoLLaMA3}, and NVILA-8B~\cite{nvila}. Our primary metric is class-averaged Top-1 accuracy ($\text{Top-1}^{\mathrm{macro}}$). For videos processed in multiple temporal windows, a video-level prediction is counted as correct if the ground-truth label is the Top-1 class in at least one window. Formally,

\begin{equation}
\text{Top-1}^{\mathrm{macro}} = \frac{1}{C} \sum_{c \in \mathcal{C}} \frac{1}{n_c} \sum_{i \in \mathcal{I}_c} \mathbf{1}\!\left\{ \exists\,k:\; \hat{\ell}^{(1)}_{i,k} = y_i \right\},
\end{equation}

where $\mathcal{C}$ is the set of classes, $C=|\mathcal{C}|$, $\mathcal{I}_c=\{i\mid y_i=c\}$, $n_c$ is the number of videos in class $c$, and $k$ indexes temporal windows. All experiments run on NVIDIA A40 or L40 GPUs (up to 46\,GB VRAM). To stabilize generation, we use conservative decoding: temperature 0.05–0.1, a cap of 64–128 new tokens, and a repetition penalty of 1.5. We use \texttt{facebook/bart-large-mnli}~\cite{BART} as a frozen NLI text classifier for scoring. Each experiment is run once (single pass) due to computational cost.
\section{Results}
\subsection{Impact of prompting across models}
We study how prompt design shapes zero-shot anomalous action recognition by evaluating Gemma-3 (4B), NVILA-8B, Qwen-2.5-VL-7B-Instruct, and VideoLLaMA-3-7B on UCF-Crime under three regimes: an unguided prompt, a guided prompt, and a guided prompt with few-shot examples (see Appendix~\ref{app:prompts} for prompts and Figure~\ref{fig:fewshot} in Appendix~\ref{app:figures} for the few-shot images). All few-shot images and descriptions are sourced from the official training split to avoid test leakage. Figure~\ref{fig:fewshot_results} reports Top-1 accuracies for the classes included in the few-shot prompts. On average, few-shot examples improve accuracy but tend to increase the false-positive rate, with Gemma-3 and NVILA benefiting the most; see Tables~\ref{tab:ucf_unguided}, \ref{tab:ucf_guided}, and \ref{tab:ucf_fewshot} in Appendix~\ref{app:results}.
\subsection{Privacy filters}
We assess robustness on RWF-2000 under privacy-preserving filters that remove personally identifiable appearance cues while retaining action-relevant structure. We consider three filters: (i) local head/face blur, where detected head regions are Gaussian-blurred using a merged mask; (ii) GAN-based anonymization from DeepPrivacy2~\cite{DeepPrivacy2} applied at the face level; and (iii) the same GAN-based anonymization extended to full-body masks. For each filter, we pre-generate a separate dataset so multiple models can be tested without reapplying the transform. All evaluations use the guided prompt; sampling, aggregation, and scoring mirror the prompting study so that differences can be attributed to privacy transforms rather than prompting or preprocessing. See results in Table~\ref{tab:rwf2000_privacy_delta} and additional details in Appendix~\ref{app:results}.

\begin{table}[b]
\centering
\begin{tabular}{lcccc}
\toprule
Model       & None & Blur Face & GAN Face & GAN Full Body \\
            & Acc (\%) / FP (\%) & $\Delta$Acc / $\Delta$FP & $\Delta$Acc / $\Delta$FP & $\Delta$Acc / $\Delta$FP \\
\midrule
Gemma-3 (4B)   & 86.25 / 20.50 & –5.0 / +10.5 & –2.8 / +7.0 & –4.0 / +7.0 \\
NVILA-8B       & 82.50 / 14.00 & –1.8 / +2.0  & –1.8 / +5.0 & –11.3 / +7.5 \\
Qwen-2.5-VL-7B-Instruct & 82.25 / 24.50 & –4.8 / +9.0  & –1.0 / +2.0 & –6.5 / +11.0 \\
VideoLLaMA-3-7B & 83.25 / 8.50  & –2.5 / +2.0  & –4.5 / –5.5 & –8.8 / –6.5 \\
\bottomrule
\end{tabular}
\caption{Baseline Top-1 accuracy and false-positive (FP) rate with no filter, and relative changes ($\Delta$, percentage points) under privacy filters on RWF-2000. Accuracy generally drops by 2–11\,pp with privacy, while FP rates often rise. VideoLLaMA-3 shows FP reductions under GAN filters.}
\label{tab:rwf2000_privacy_delta}
\end{table}


\section{Conclusions}
\vspace{-0.1 cm}
This work evaluated small vision-LLMs for zero-shot anomaly detection, revealing a critical trade-off between prompting techniques, privacy filters and accuracy. While few-shot prompting improves accuracy for some models, it often increases false-positive rates. Privacy-preserving filters, crucial for deployment, induce a modest performance drop, with full-body GAN anonymization being the most disruptive due to video inconsistencies.
Overall, these models show promise for simple tasks like fight detection but are not yet reliable enough for complex, autonomous surveillance. Future efforts must focus on improving the temporal consistency of privacy methods and balancing model sensitivity with precision.

\newpage
\bibliographystyle{plainnat}
\bibliography{bibliography}

\begin{thebibliography}{18}
\providecommand{\natexlab}[1]{#1}
\providecommand{\url}[1]{\texttt{#1}}
\expandafter\ifx\csname urlstyle\endcsname\relax
  \providecommand{\doi}[1]{doi: #1}\else
  \providecommand{\doi}{doi: \begingroup \urlstyle{rm}\Url}\fi

\bibitem[Bai et~al.(2025)Bai, Chen, Liu, Wang, Ge, Song, Dang, Wang, Wang, Tang, Zhong, Zhu, Yang, Li, Wan, Wang, Ding, Fu, Xu, Ye, Zhang, Xie, Cheng, Zhang, Yang, Xu, and Lin]{Qwen2.5-VL}
Shuai Bai, Keqin Chen, Xuejing Liu, Jialin Wang, Wenbin Ge, Sibo Song, Kai Dang, Peng Wang, Shijie Wang, Jun Tang, Humen Zhong, Yuanzhi Zhu, Mingkun Yang, Zhaohai Li, Jianqiang Wan, Pengfei Wang, Wei Ding, Zheren Fu, Yiheng Xu, Jiabo Ye, Xi~Zhang, Tianbao Xie, Zesen Cheng, Hang Zhang, Zhibo Yang, Haiyang Xu, and Junyang Lin.
\newblock Qwen2.5-vl technical report.
\newblock \emph{arXiv preprint arXiv:2502.13923}, 2025.

\bibitem[Chen et~al.(2023)Chen, Ma, Jian~Yew, Hur, and Khoo]{tevad}
Weiling Chen, Keng~Teck Ma, Zi~Jian~Yew, Minhoe Hur, and David Aik-Aun Khoo.
\newblock Tevad: Improved video anomaly detection with captions.
\newblock In \emph{2023 IEEE/CVF Conference on Computer Vision and Pattern Recognition Workshops (CVPRW)}, pages 5549--5559, 2023.
\newblock \doi{10.1109/CVPRW59228.2023.00587}.

\bibitem[Cheng et~al.(2021)Cheng, Cai, and Li]{RWF2000}
Ming Cheng, Kunjing Cai, and Ming Li.
\newblock Rwf-2000: An open large scale video database for violence detection.
\newblock In \emph{2020 25th International Conference on Pattern Recognition (ICPR)}, pages 4183--4190, 2021.
\newblock \doi{10.1109/ICPR48806.2021.9412502}.

\bibitem[Hukkelås and Lindseth(2023)]{DeepPrivacy2}
Håkon Hukkelås and Frank Lindseth.
\newblock Deepprivacy2: Towards realistic full-body anonymization.
\newblock In \emph{2023 IEEE/CVF Winter Conference on Applications of Computer Vision (WACV)}, pages 1329--1338, 2023.
\newblock \doi{10.1109/WACV56688.2023.00138}.

\bibitem[Karim et~al.(2024)Karim, Doshi, and Yilmaz]{REWARD}
Hamza Karim, Keval Doshi, and Yasin Yilmaz.
\newblock Real-time weakly supervised video anomaly detection.
\newblock In \emph{Proceedings of the IEEE/CVF Winter Conference on Applications of Computer Vision (WACV)}, pages 6848--6856, January 2024.

\bibitem[Lewis et~al.(2019)Lewis, Liu, Goyal, Ghazvininejad, Mohamed, Levy, Stoyanov, and Zettlemoyer]{BART}
Mike Lewis, Yinhan Liu, Naman Goyal, Marjan Ghazvininejad, Abdelrahman Mohamed, Omer Levy, Veselin Stoyanov, and Luke Zettlemoyer.
\newblock {BART:} denoising sequence-to-sequence pre-training for natural language generation, translation, and comprehension.
\newblock \emph{CoRR}, abs/1910.13461, 2019.
\newblock URL \url{http://arxiv.org/abs/1910.13461}.

\bibitem[Liu et~al.(2024)Liu, Zhu, Shi, Zhang, Lou, Yang, Xi, Cao, Gu, Li, Li, Fang, Chen, Hsieh, Huang, Cheng, Nath, Hu, Liu, Krishna, Xu, Wang, Molchanov, Kautz, Yin, Han, and Lu]{nvila}
Zhijian Liu, Ligeng Zhu, Baifeng Shi, Zhuoyang Zhang, Yuming Lou, Shang Yang, Haocheng Xi, Shiyi Cao, Yuxian Gu, Dacheng Li, Xiuyu Li, Yunhao Fang, Yukang Chen, Cheng-Yu Hsieh, De-An Huang, An-Chieh Cheng, Vishwesh Nath, Jinyi Hu, Sifei Liu, Ranjay Krishna, Daguang Xu, Xiaolong Wang, Pavlo Molchanov, Jan Kautz, Hongxu Yin, Song Han, and Yao Lu.
\newblock Nvila: Efficient frontier visual language models, 2024.
\newblock URL \url{https://arxiv.org/abs/2412.04468}.

\bibitem[Meo et~al.(2024{\natexlab{a}})Meo, Lica, Ikram, Nakano, Shah, Didolkar, Liu, Goyal, and Dauwels]{meo2024masked}
Cristian Meo, Mircea Lica, Zarif Ikram, Akihiro Nakano, Vedant Shah, Aniket~Rajiv Didolkar, Dianbo Liu, Anirudh Goyal, and Justin Dauwels.
\newblock Masked generative priors improve world models sequence modelling capabilities.
\newblock \emph{arXiv preprint arXiv:2410.07836}, 2024{\natexlab{a}}.

\bibitem[Meo et~al.(2024{\natexlab{b}})Meo, Nakano, Lic{\u{a}}, Didolkar, Suzuki, Goyal, Zhang, Dauwels, Matsuo, and Bengio]{meo2024object}
Cristian Meo, Akihiro Nakano, Mircea Lic{\u{a}}, Aniket Didolkar, Masahiro Suzuki, Anirudh Goyal, Mengmi Zhang, Justin Dauwels, Yutaka Matsuo, and Yoshua Bengio.
\newblock Object-centric temporal consistency via conditional autoregressive inductive biases.
\newblock \emph{arXiv preprint arXiv:2410.15728}, 2024{\natexlab{b}}.

\bibitem[Sultani et~al.(2018)Sultani, Chen, and Shah]{UCFCrime}
Waqas Sultani, Chen Chen, and Mubarak Shah.
\newblock Real-world anomaly detection in surveillance videos.
\newblock \emph{CoRR}, abs/1801.04264, 2018.
\newblock URL \url{http://arxiv.org/abs/1801.04264}.

\bibitem[Team et~al.(2025)Team, Kamath, Ferret, Pathak, Vieillard, Merhej, Perrin, Matejovicova, Ramé, Rivière, Rouillard, Mesnard, Cideron, bastien Grill, Ramos, Yvinec, Casbon, Pot, Penchev, Liu, Visin, Kenealy, Beyer, Zhai, Tsitsulin, Busa-Fekete, Feng, Sachdeva, Coleman, Gao, Mustafa, Barr, Parisotto, Tian, Eyal, Cherry, Peter, Sinopalnikov, Bhupatiraju, Agarwal, Kazemi, Malkin, Kumar, Vilar, Brusilovsky, Luo, Steiner, Friesen, Sharma, Sharma, Gilady, Goedeckemeyer, Saade, Feng, Kolesnikov, Bendebury, Abdagic, Vadi, György, Pinto, Das, Bapna, Miech, Yang, Paterson, Shenoy, Chakrabarti, Piot, Wu, Shahriari, Petrini, Chen, Lan, Choquette-Choo, Carey, Brick, Deutsch, Eisenbud, Cattle, Cheng, Paparas, Sreepathihalli, Reid, Tran, Zelle, Noland, Huizenga, Kharitonov, Liu, Amirkhanyan, Cameron, Hashemi, Klimczak-Plucińska, Singh, Mehta, Lehri, Hazimeh, Ballantyne, Szpektor, Nardini, Pouget-Abadie, Chan, Stanton, Wieting, Lai, Orbay, Fernandez, Newlan, yeong Ji, Singh, Black, Yu, Hui, Vodrahalli, Greff, Qiu,
  Valentine, Coelho, Ritter, Hoffman, Watson, Chaturvedi, Moynihan, Ma, Babar, Noy, Byrd, Roy, Momchev, Chauhan, Sachdeva, Bunyan, Botarda, Caron, Rubenstein, Culliton, Schmid, Sessa, Xu, Stanczyk, Tafti, Shivanna, Wu, Pan, Rokni, Willoughby, Vallu, Mullins, Jerome, Smoot, Girgin, Iqbal, Reddy, Sheth, Põder, Bhatnagar, Panyam, Eiger, Zhang, Liu, Yacovone, Liechty, Kalra, Evci, Misra, Roseberry, Feinberg, Kolesnikov, Han, Kwon, Chen, Chow, Zhu, Wei, Egyed, Cotruta, Giang, Kirk, Rao, Black, Babar, Lo, Moreira, Martins, Sanseviero, Gonzalez, Gleicher, Warkentin, Mirrokni, Senter, Collins, Barral, Ghahramani, Hadsell, Matias, Sculley, Petrov, Fiedel, Shazeer, Vinyals, Dean, Hassabis, Kavukcuoglu, Farabet, Buchatskaya, Alayrac, Anil, Dmitry, Lepikhin, Borgeaud, Bachem, Joulin, Andreev, Hardin, Dadashi, and Hussenot]{gemma3}
Gemma Team, Aishwarya Kamath, Johan Ferret, Shreya Pathak, Nino Vieillard, Ramona Merhej, Sarah Perrin, Tatiana Matejovicova, Alexandre Ramé, Morgane Rivière, Louis Rouillard, Thomas Mesnard, Geoffrey Cideron, Jean bastien Grill, Sabela Ramos, Edouard Yvinec, Michelle Casbon, Etienne Pot, Ivo Penchev, Gaël Liu, Francesco Visin, Kathleen Kenealy, Lucas Beyer, Xiaohai Zhai, Anton Tsitsulin, Robert Busa-Fekete, Alex Feng, Noveen Sachdeva, Benjamin Coleman, Yi~Gao, Basil Mustafa, Iain Barr, Emilio Parisotto, David Tian, Matan Eyal, Colin Cherry, Jan-Thorsten Peter, Danila Sinopalnikov, Surya Bhupatiraju, Rishabh Agarwal, Mehran Kazemi, Dan Malkin, Ravin Kumar, David Vilar, Idan Brusilovsky, Jiaming Luo, Andreas Steiner, Abe Friesen, Abhanshu Sharma, Abheesht Sharma, Adi~Mayrav Gilady, Adrian Goedeckemeyer, Alaa Saade, Alex Feng, Alexander Kolesnikov, Alexei Bendebury, Alvin Abdagic, Amit Vadi, András György, André~Susano Pinto, Anil Das, Ankur Bapna, Antoine Miech, Antoine Yang, Antonia Paterson, Ashish
  Shenoy, Ayan Chakrabarti, Bilal Piot, Bo~Wu, Bobak Shahriari, Bryce Petrini, Charlie Chen, Charline~Le Lan, Christopher~A. Choquette-Choo, CJ~Carey, Cormac Brick, Daniel Deutsch, Danielle Eisenbud, Dee Cattle, Derek Cheng, Dimitris Paparas, Divyashree~Shivakumar Sreepathihalli, Doug Reid, Dustin Tran, Dustin Zelle, Eric Noland, Erwin Huizenga, Eugene Kharitonov, Frederick Liu, Gagik Amirkhanyan, Glenn Cameron, Hadi Hashemi, Hanna Klimczak-Plucińska, Harman Singh, Harsh Mehta, Harshal~Tushar Lehri, Hussein Hazimeh, Ian Ballantyne, Idan Szpektor, Ivan Nardini, Jean Pouget-Abadie, Jetha Chan, Joe Stanton, John Wieting, Jonathan Lai, Jordi Orbay, Joseph Fernandez, Josh Newlan, Ju~yeong Ji, Jyotinder Singh, Kat Black, Kathy Yu, Kevin Hui, Kiran Vodrahalli, Klaus Greff, Linhai Qiu, Marcella Valentine, Marina Coelho, Marvin Ritter, Matt Hoffman, Matthew Watson, Mayank Chaturvedi, Michael Moynihan, Min Ma, Nabila Babar, Natasha Noy, Nathan Byrd, Nick Roy, Nikola Momchev, Nilay Chauhan, Noveen Sachdeva, Oskar
  Bunyan, Pankil Botarda, Paul Caron, Paul~Kishan Rubenstein, Phil Culliton, Philipp Schmid, Pier~Giuseppe Sessa, Pingmei Xu, Piotr Stanczyk, Pouya Tafti, Rakesh Shivanna, Renjie Wu, Renke Pan, Reza Rokni, Rob Willoughby, Rohith Vallu, Ryan Mullins, Sammy Jerome, Sara Smoot, Sertan Girgin, Shariq Iqbal, Shashir Reddy, Shruti Sheth, Siim Põder, Sijal Bhatnagar, Sindhu~Raghuram Panyam, Sivan Eiger, Susan Zhang, Tianqi Liu, Trevor Yacovone, Tyler Liechty, Uday Kalra, Utku Evci, Vedant Misra, Vincent Roseberry, Vlad Feinberg, Vlad Kolesnikov, Woohyun Han, Woosuk Kwon, Xi~Chen, Yinlam Chow, Yuvein Zhu, Zichuan Wei, Zoltan Egyed, Victor Cotruta, Minh Giang, Phoebe Kirk, Anand Rao, Kat Black, Nabila Babar, Jessica Lo, Erica Moreira, Luiz~Gustavo Martins, Omar Sanseviero, Lucas Gonzalez, Zach Gleicher, Tris Warkentin, Vahab Mirrokni, Evan Senter, Eli Collins, Joelle Barral, Zoubin Ghahramani, Raia Hadsell, Yossi Matias, D.~Sculley, Slav Petrov, Noah Fiedel, Noam Shazeer, Oriol Vinyals, Jeff Dean, Demis Hassabis,
  Koray Kavukcuoglu, Clement Farabet, Elena Buchatskaya, Jean-Baptiste Alayrac, Rohan Anil, Dmitry, Lepikhin, Sebastian Borgeaud, Olivier Bachem, Armand Joulin, Alek Andreev, Cassidy Hardin, Robert Dadashi, and Léonard Hussenot.
\newblock Gemma 3 technical report, 2025.
\newblock URL \url{https://arxiv.org/abs/2503.19786}.

\bibitem[Wu et~al.(2020)Wu, Liu, Shi, Sun, Shao, Wu, and Yang]{XD_violence}
Peng Wu, jing Liu, Yujia Shi, Yujia Sun, Fangtao Shao, Zhaoyang Wu, and Zhiwei Yang.
\newblock Not only look, but also listen: Learning multimodal violence detection under weak supervision.
\newblock In \emph{European Conference on Computer Vision (ECCV)}, 2020.

\bibitem[Wu et~al.(2024)Wu, Zhou, Pang, Sun, Liu, Wang, and Zhang]{openvocab}
Peng Wu, Xuerong Zhou, Guansong Pang, Yujia Sun, Jing Liu, Peng Wang, and Yanning Zhang.
\newblock Open-vocabulary video anomaly detection.
\newblock In \emph{Proceedings of the IEEE/CVF Conference on Computer Vision and Pattern Recognition}, pages 18297--18307, 2024.

\bibitem[Yun et~al.(2025)Yun, Masukawa, Na, and Imani]{misionGNN}
Sanggeon Yun, Ryozo Masukawa, Minhyoung Na, and Mohsen Imani.
\newblock Missiongnn: Hierarchical multimodal gnn-based weakly supervised video anomaly recognition with mission-specific knowledge graph generation.
\newblock In \emph{2025 IEEE/CVF Winter Conference on Applications of Computer Vision (WACV)}, pages 4736--4745. IEEE, 2025.

\bibitem[Zanella et~al.(2024{\natexlab{a}})Zanella, Liberatori, Menapace, Poiesi, Wang, and Ricci]{AnomalyCLIP}
Luca Zanella, Benedetta Liberatori, Willi Menapace, Fabio Poiesi, Yiming Wang, and Elisa Ricci.
\newblock Delving into clip latent space for video anomaly recognition.
\newblock \emph{Comput. Vis. Image Underst.}, 249:\penalty0 104163, 2024{\natexlab{a}}.
\newblock URL \url{https://doi.org/10.1016/j.cviu.2024.104163}.

\bibitem[Zanella et~al.(2024{\natexlab{b}})Zanella, Menapace, Mancini, Wang, and Ricci]{lavad}
Luca Zanella, Willi Menapace, Massimiliano Mancini, Yiming Wang, and Elisa Ricci.
\newblock Harnessing large language models for training-free video anomaly detection.
\newblock In \emph{Proceedings of the IEEE/CVF Conference on Computer Vision and Pattern Recognition}, pages 18527--18536, 2024{\natexlab{b}}.

\bibitem[Zhang et~al.(2025)Zhang, Li, Cheng, Hu, Yuan, Chen, Leng, Jiang, Zhang, Li, Jin, Zhang, Wang, Bing, and Zhao]{VideoLLaMA3}
Boqiang Zhang, Kehan Li, Zesen Cheng, Zhiqiang Hu, Yuqian Yuan, Guanzheng Chen, Sicong Leng, Yuming Jiang, Hang Zhang, Xin Li, Peng Jin, Wenqi Zhang, Fan Wang, Lidong Bing, and Deli Zhao.
\newblock Videollama 3: Frontier multimodal foundation models for image and video understanding.
\newblock \emph{arXiv preprint arXiv:2501.13106}, 2025.
\newblock URL \url{https://arxiv.org/abs/2501.13106}.

\bibitem[Zhang et~al.(2024)Zhang, Xu, Wang, Zuo, Han, Huang, Gao, Wang, and Sang]{holmesVAD}
Huaxin Zhang, Xiaohao Xu, Xiang Wang, Jialong Zuo, Chuchu Han, Xiaonan Huang, Changxin Gao, Yuehuan Wang, and Nong Sang.
\newblock Holmes-vad: Towards unbiased and explainable video anomaly detection via multi-modal llm.
\newblock \emph{arXiv preprint arXiv:2406.12235}, 2024.

\end{thebibliography}

\appendix

\newpage
\section{Prompts used in experiments} \label{app:prompts}

\begin{lstlisting}[breaklines=true, basicstyle=\ttfamily]
UNGUIDED_PROMPT_UCF = """
    You are given a short video clip. Identify the primary action(s) you see and provide a concise description (<= 40 words).
"""
\end{lstlisting}

\begin{lstlisting}[breaklines=true, basicstyle=\ttfamily]
GUIDED_PROMPT = """
    You are given a short video clip. Analyze it and respond in the following format:
    
    [Predicted Class]: Brief description of actions happening in the input frames (<= 40 words).
    
    Choose the most likely class from the options below.
    
    1. Abuse: Person being abused or assaulted by another individual.
    2. Arrest: Law enforcement detaining or arresting individuals.
    3. Arson: Deliberate setting of fire causing a blaze.
    4. Assault: Physical attack (punching, kicking, hitting).
    5. Burglary: Unauthorized intrusion to commit theft.
    6. Explosion: Sudden blast or large fireball.
    7. Fighting: Close-quarters physical fight (wrestling, brawling).
    8. Normal: Routine, non-violent, everyday activity.
    9. RoadAccidents: Vehicle collision or traffic accident.
    10. Robbery: Theft involving force or threat from a person.
    11. Shooting: Discharge of a firearm (gun visible or muzzle flash).
    12. Shoplifting: Theft from a store without force or threat.
    13. Stealing: Theft of objects without direct confrontation.
    14. Vandalism: Deliberate damage or destruction of property.
"""
\end{lstlisting

\begin{lstlisting}[breaklines=true, basicstyle=\ttfamily]
FEW_SHOT_EXAMPLES: List[Dict[str, Any]] = [
    {"role": "user", "content": [{"type": "image_url", "image_url": {"url": "demo_images/few_shot/Shooting.png"}}]},
    {"role": "assistant", "content": "A person with raised arm firing a gun as seen from the muzzle flash. Label: Shooting."},

    {"role": "user", "content": [{"type": "image_url", "image_url": {"url": "demo_images/few_shot/RoadAccidents.png"}}]},
    {"role": "assistant", "content": "A car crashes seen from the smoke on the right. Label: RoadAccidents."},

    {"role": "user", "content": [{"type": "image_url", "image_url": {"url": "demo_images/few_shot/Fighting.png"}}]},
    {"role": "assistant", "content": "Two persons trying to hit people. Label: Fighting."},

    {"role": "user", "content": [{"type": "image_url", "image_url": {"url": "demo_images/few_shot/Stealing.png"}}]},
    {"role": "assistant", "content": "A person breaking into a car. Label: Stealing."},
]
\end{lstlisting}

\begin{lstlisting}[breaklines=true, basicstyle=\ttfamily]
GUIDED_PROMPT_RWF2000 = """
    You are given a short surveillance video clip. Analyze it and respond in the following format:
    
    [Predicted Class]: Brief description of actions happening in the input frames (<= 40 words).
    
    Choose the most likely class from the options below.
    
    1. Fighting: Physical altercation between individuals (e.g., punching, pushing, brawling).
    2. Normal: Routine, peaceful activities with no signs of aggression or conflict.
"""
\end{lstlisting}

\section{Additional figures} \label{app:figures}
Few-Shot prompting images
\begin{figure}[h]
    \centering
    \includegraphics[width=1.0\linewidth]{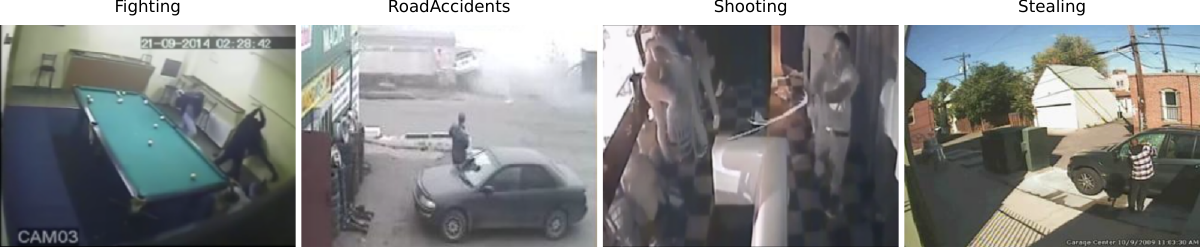}
    \caption{Images from UCF-Crime dataset used for few-shot prompting}
    \label{fig:fewshot}
\end{figure}

\begin{figure}[t]
  \centering
  \begin{subfigure}[t]{0.467\linewidth}
    \centering
    \includegraphics[width=\linewidth]{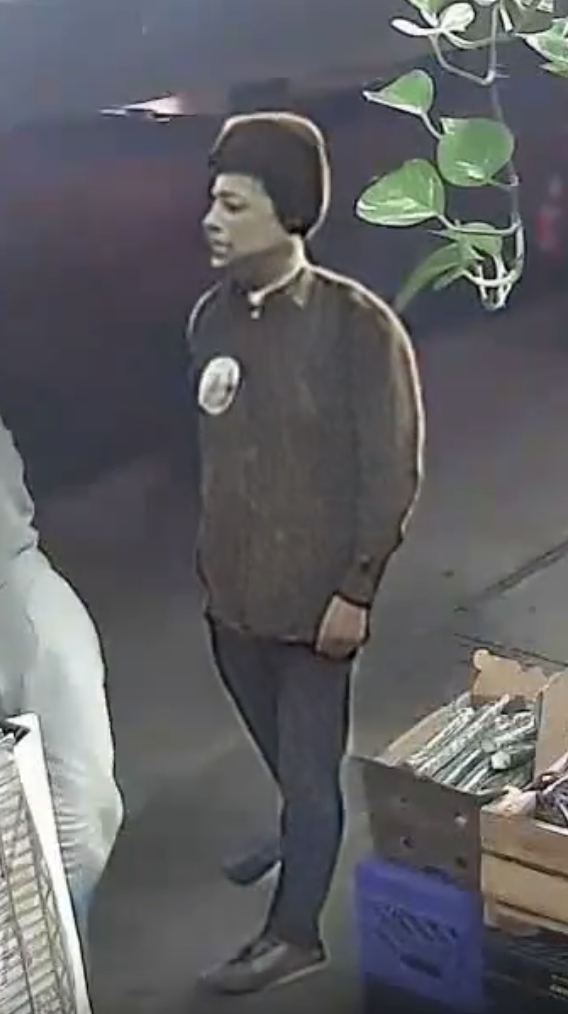}
    \caption{Example A}
    \label{fig:first}
  \end{subfigure}
  \hfill
  \begin{subfigure}[t]{0.433\linewidth}
    \centering
    \includegraphics[width=\linewidth]{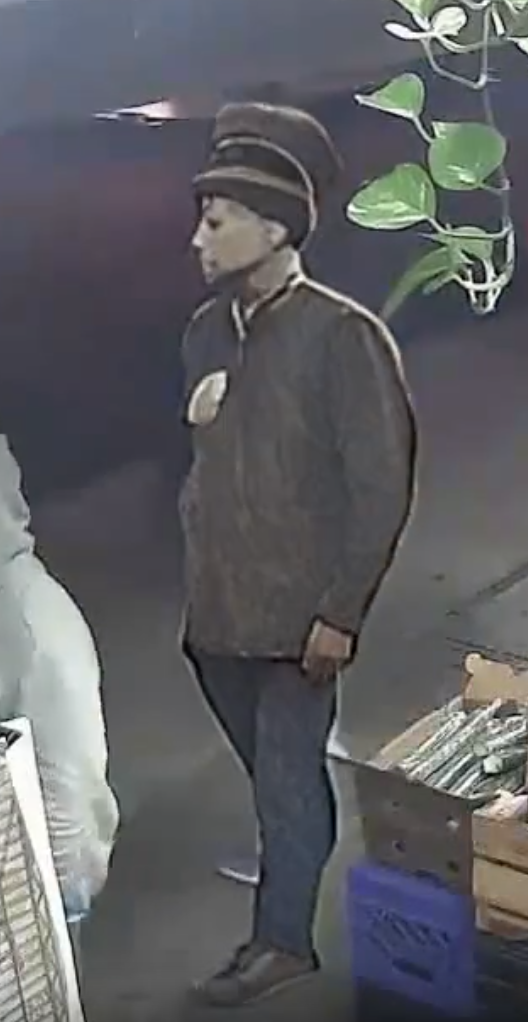}
    \caption{Example B}
    \label{fig:second}
  \end{subfigure}
  \caption{Two examples of the same person generated slightly differently by the GAN, leading to inconsistent motion in video.}
  \label{fig:both}
\end{figure}

\section{Experimental results} \label{app:results}

\begin{figure}
    \includegraphics[width=1.2\linewidth]{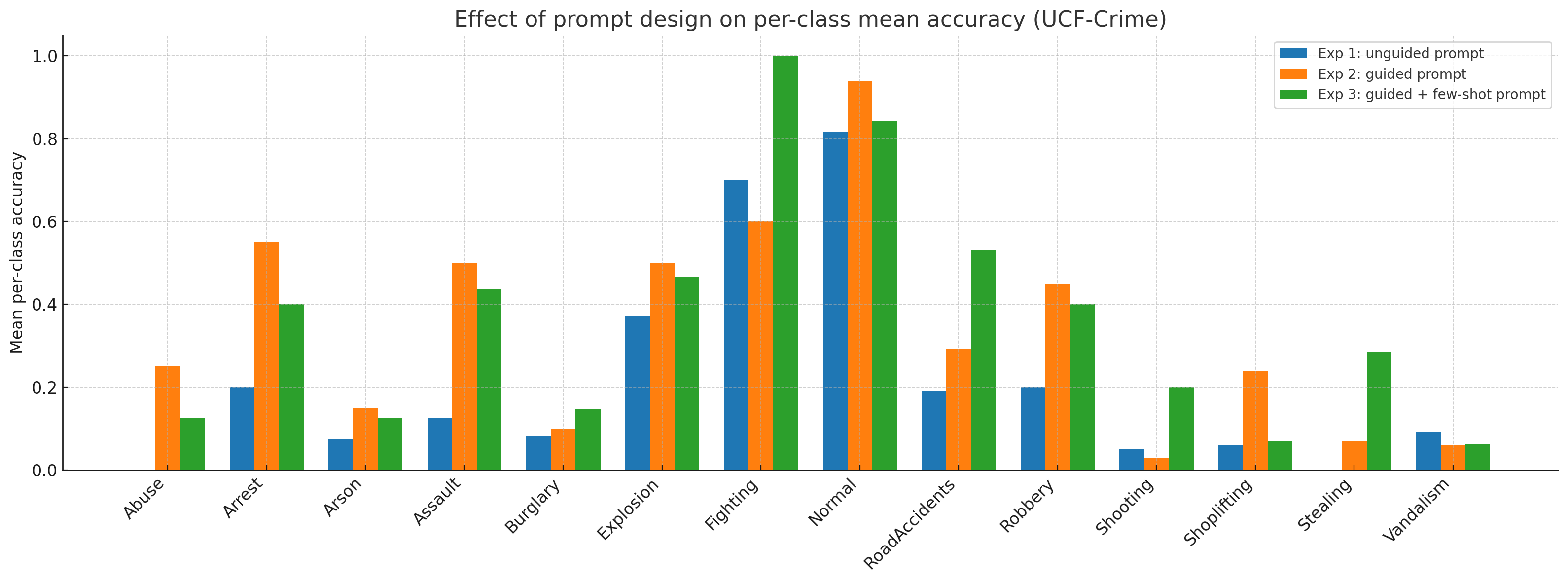}
    \caption{All classes compared over prompting experiments, the few-shot examples include: Fighting, RoadAccidents, Shooting and Stealing.}
    \label{fig:prompt_experiment_bar}
\end{figure}

Figure~\ref{fig:prompt_experiment_bar} contains the results of all classes compared over the prompting experiments, the tables below show results for each individual experiment. AUC is taken as the batch (256 frames) level score with each class other than "Normal" labeled as anomaly. FP shows the percentage of batches predicted as an other class than "Normal" in videos that are labeled "Normal". Wrong label indicates a label being present in the generated text which does not correspond to the video label. All experiments on RFW2000 share the guided prompt.

\begin{table}[]
    \centering
    \caption{UCF-Crime (Unguided Prompt)}
    \label{tab:ucf_unguided}
    \begin{tabular}{lcccc}
    \toprule
    Model & Top-1 (\%) & AUC (\%) & FP (\%) & Wrong Label (\%) \\
    \midrule
    Gemma3-4B & 26.29 & 65.68 & 11.33 & 4.56 \\
    NVILA-8B  & 13.39 & 56.96 & 6.67  & 0.39 \\
    Qwen2.5   & 25.31 & 64.14 & 11.00 & 2.21 \\
    VideoLLama3 & 19.94 & 50.05 & 80.67 & 4.17 \\
    \bottomrule
    \end{tabular}
\end{table}

\begin{table}[]
    \centering
    \caption{UCF-Crime (Guided Prompt)}
    \label{tab:ucf_guided}
    \begin{tabular}{lcccc}
    \toprule
    Model & Top-1 (\%) & AUC (\%) & FP (\%) & Wrong Label (\%) \\
    \midrule
    Gemma3-4B & 33.85 & 77.71 & 21.67  & 58.20 \\
    NVILA-8B  & 27.00 & 78.97 & 5.00  & 56.38 \\
    Qwen2.5   & 34.69 & 74.63 & 10.67 & 76.56 \\
    VideoLLama3 & 34.16 & 73.40 & 19.67 & 42.19 \\
    \bottomrule
    \end{tabular}
\end{table}

\begin{table}[]
    \centering
    \caption{UCF-Crime (Guided Prompt + Few-Shot Examples)}
    \label{tab:ucf_fewshot}
    \begin{tabular}{lcccc}
    \toprule
    Model & Top-1 (\%) & AUC (\%) & FP (\%) & Wrong Label (\%) \\
    \midrule
    Gemma3-4B & 29.80 & 57.73 & 68.67  & 42.97 \\
    NVILA-8B  & 45.05 & 67.06 & 18.00 & 47.79 \\
    Qwen2.5   & 38.87 & 75.22 & 9.33  & 73.24 \\
    VideoLLama3 & 31.44 & 69.61 & 5.00 & 47.27 \\
    \bottomrule
    \end{tabular}
\end{table}

\begin{table}[]
    \centering
    \caption{UCF-Crime (Guided Prompt + Privacy Filter)}
    \begin{tabular}{lcccc}
    \toprule
    Model & Top-1 (\%) & AUC (\%) & FP (\%) & Wrong Label (\%) \\
    \midrule
    Gemma3-4B & 34.33 & 75.19 & 29.33  & 61.72 \\
    NVILA-8B  & 28.14 & 77.22 & 9.33  & 58.20 \\
    Qwen2.5   & 34.62 & 75.70 & 14.67 & 76.56 \\
    VideoLLama3 & 27.74 & 68.95 & 34.00 & 39.19 \\
    \bottomrule
    \end{tabular}
\end{table}

\subsection{RWF2000 experiments}

\begin{table}[]
    \centering
    \caption{RWF2000}
    \begin{tabular}{lccc}
    \toprule
    Model & Top-1 (\%) & FP (\%) & Wrong Label (\%) \\
    \midrule
    Gemma3-4B & 86.25 & 20.50 & 16.75 \\
    NVILA-8B  & 82.50 & 14.00 & 54.50 \\
    Qwen2.5   & 82.25 & 24.50 & 88.50 \\
    VideoLLama3 & 83.25 & 8.50 & 14.25 \\
    \bottomrule
    \end{tabular}
\end{table}

\begin{table}[]
    \centering
    \caption{RWF2000 (With privacy filter - blur face)}
    \begin{tabular}{lccc}
    \toprule
    Model & Top-1 (\%) & FP (\%) & Wrong Label (\%) \\
    \midrule
    Gemma3-4B & 81.25 & 31.00 & 21.25 \\
    NVILA-8B  & 80.75 & 16.00 & 56.25 \\
    Qwen2.5   & 77.50 & 33.50 & 92.50 \\
    VideoLLama3 & 80.75 & 10.50 & 17.75 \\
    \bottomrule
    \end{tabular}
\end{table}

\begin{table}[]
    \centering
    \caption{RWF2000 (With privacy filter - GAN face)}
    \begin{tabular}{lccc}
    \toprule
    Model & Top-1 (\%) & FP (\%) & Wrong Label (\%) \\
    \midrule
    Gemma3-4B & 83.50 & 27.50 & 19.25 \\
    NVILA-8B  & 80.75 & 19.00 & 58.00 \\
    Qwen2.5   & 81.25 & 26.50 & 91.50 \\
    VideoLLama3 & 78.75 & 3.00 & 24.00 \\
    \bottomrule
    \end{tabular}
\end{table}

\begin{table}[]
    \centering
    \caption{RWF2000 (With privacy filter - GAN full body)}
    \begin{tabular}{lccc}
    \toprule
    Model & Top-1 (\%) & FP (\%) & Wrong Label (\%) \\
    \midrule
    Gemma3-4B & 82.25 & 27.50 & 23.75 \\
    NVILA-8B  & 73.25 & 21.50 & 59.75 \\
    Qwen2.5   & 75.75 & 35.50 & 95.50 \\
    VideoLLama3 & 74.50 & 2.00 & 27.25 \\
    \bottomrule
    \end{tabular}
\end{table}

\end{document}